\documentclass[conference]{IEEEtran}
\usepackage{cite}
\usepackage{amsmath,amssymb,amsfonts}
\usepackage{algorithmic}
\usepackage{graphicx}
\usepackage{textcomp}
\usepackage{xcolor}
\usepackage{url}
\usepackage{hyperref}
\def\BibTeX{{\rm B\kern-.05em{\sc i\kern-.025em b}\kern-.08em
    T\kern-.1667em\lower.7ex\hbox{E}\kern-.125emX}}
\begin{document}

\title{FACTS: Fine-Grained Action Classification for Tactical Sports}

\author{
\IEEEauthorblockN{Christopher Lai\textsuperscript{1}, Jason Mo\textsuperscript{2}, Haotian Xia\textsuperscript{3}, Yuan-fang Wang\textsuperscript{1}}
\IEEEauthorblockA{\textsuperscript{1}\textit{University of California, Santa Barbara}, \textsuperscript{2}\textit{Georgia Institute of Technology}, \textsuperscript{3}\textit{University of California, Irvine} \\
\textsuperscript{1}\{christopherlai, yfwang\}@ucsb.edu, \textsuperscript{2}thejasonmo@gatech.edu, \textsuperscript{3}xiah6@uci.edu}
}

\maketitle

\begin{abstract}
Classifying fine-grained actions in fast-paced, close-combat sports such as fencing and boxing presents unique challenges due to the complexity, speed, and nuance of movements. Traditional methods reliant on pose estimation or fancy sensor data often struggle to capture these dynamics accurately. We introduce FACTS, a novel transformer-based approach for fine-grained action recognition that processes raw video data directly, eliminating the need for pose estimation and the use of cumbersome body markers and sensors. FACTS achieves state-of-the-art performance, with 90\% accuracy on fencing actions and 83.25\% on boxing actions. Additionally, we present a new publicly available dataset featuring 8 detailed fencing actions, addressing critical gaps in sports analytics resources. Our findings enhance training, performance analysis, and spectator engagement, setting a new benchmark for action classification in tactical sports. The dataset is available at \href{https://anonymous.4open.science/r/FACTS-B1C5}{https://anonymous.4open.science/r/FACTS-B1C5}.
\end{abstract}


\section{Introduction and Background}
\label{sec:intro}

Recent advancements in artificial intelligence have sparked increased research in sports analytics, with a primary focus on action recognition. However, much of this work has been limited to coarse classification tasks, such as identifying a pass, a shot, or a dribble in basketball. While valuable, such general classifications fail to provide the granular insights that players, coaches, and analysts require. For example, differentiating between a jump shot and a fade-away shot, or a crossover dribble and a behind-the-back dribble can offer a deeper understanding of player techniques and strategies.

In this study, we focus on fine-grained action understanding, particularly in high-speed, close-combat sports. Fencing and boxing were chosen as sports in our study due to their demonstration of fine-grained actions in fast-paced, close-combat settings. Both of these sports require precise movements, tactical decision-making, and rapid adaptations in response to an opponent's actions. These traits make them ideal for studying detailed, high-frequency movement patterns and provide a unique challenge to traditional action recognition models.

The ability to classify nuanced actions has far-reaching applications:
\begin{itemize}
    \item \textbf{Amateurs:} Enables better learning and skill development by breaking down complex movements into digestible segments.
    \item \textbf{Athletes:} Assists in analyzing techniques, identifying patterns, and optimizing strategies.
    \item \textbf{Coaches:} Provides actionable insights for training plans, focusing on both strengths and areas of improvement.
    \item \textbf{Trainers:} Assists in early injury prevention, monitoring of recovery progress, and setting performance goals.
    \item \textbf{Sports Broadcasters:} Simplifies complex actions for audiences, enhancing viewer engagement and understanding.
\end{itemize}

Traditional methods for action recognition often rely on 3D pose estimation \cite{pavlakos2017harvesting, trumble2017total}, which depends on specialized equipment and is prone to noise, occlusion, and inaccuracies. Even with state-of-the-art pose estimation frameworks, the results often fail to meet the precision required for fine-grained action classification. Approaches using RNNs, LSTMs, and GCNs \cite{SholtoDouglas, pageaud2019sport}, rely heavily on preprocessing steps like segmentation or skeleton analysis, which can introduce additional errors and limit the models' generalizability.

\begin{figure}[htbp]
    \centerline{\includegraphics[width=0.5\textwidth]{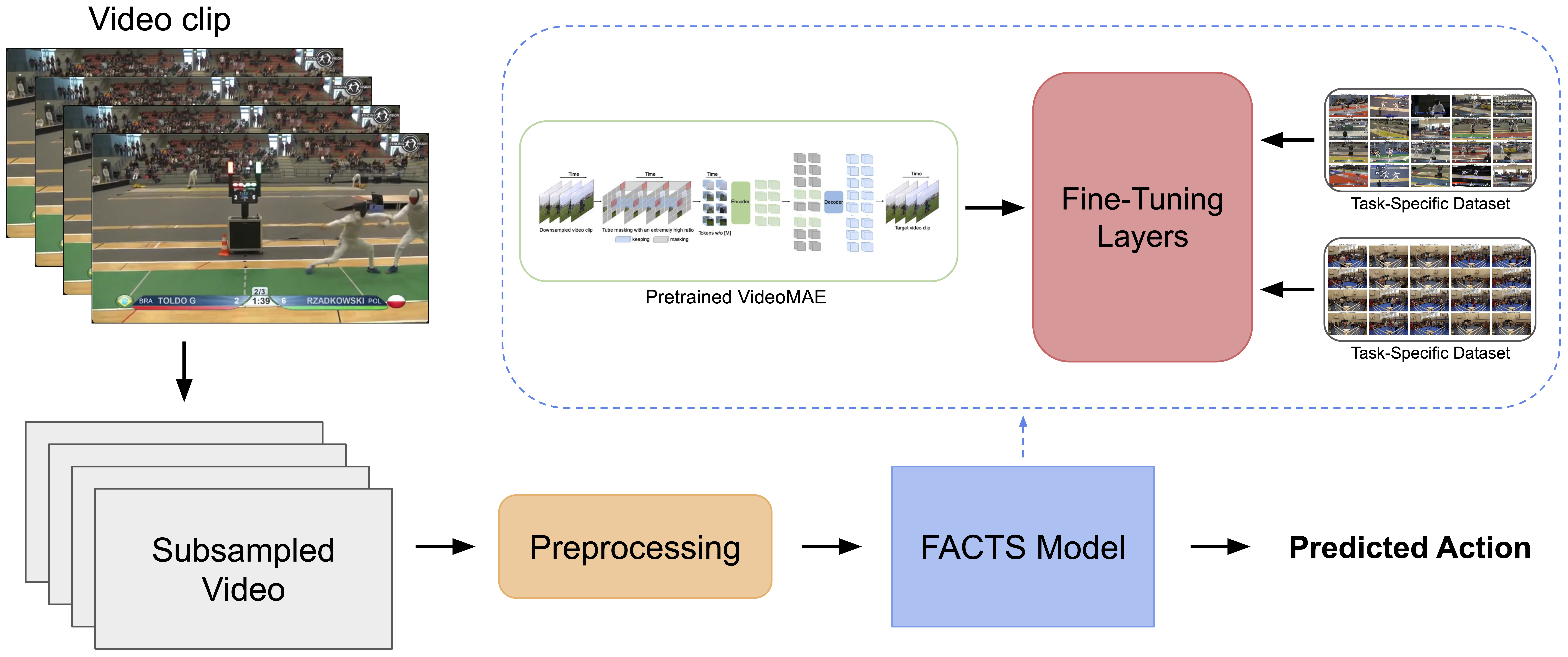}}
    \caption{Overview of the FACTS model architecture, leveraging VideoMAE's encoder-decoder framework with temporal positional encodings}
    \label{fig:model_diagram}
\end{figure}

To overcome these challenges, we present FACTS, a novel framework that directly processes raw video data, bypassing pose estimation and segmentation. By removing sport-specific tasks such as skeleton analysis and segmentation, we can better generalize across different fast-paced sports. Our approach, captured in Fig.~\ref{fig:model_diagram}, is designed to capture both spatial and temporal nuances, enabling precise classification of fine-grained actions in high-speed sports like fencing and boxing.

In summary, our contributions are threefold:
\begin{itemize}
    \item We present a novel approach for fine-grained action recognition that achieves SOTA performance with 90\% accuracy in fencing and 83.25\% in boxing, while also removing pose estimation and use of special body makers and sensors, heavily simplifying the process and reducing the cost.
    \item We introduce a dataset tailored for fine-grained action classification in tactical sports, featuring 8 distinct fencing actions such as attack, riposte, counterattack, and remise.
    \item We will make the FACTS dataset publicly available to support advancements in the broader field of action recognition in complex, fast-paced sports.
\end{itemize}

\section{Related Work}
\label{sec:related}

\subsection{AI Referees in Fencing}
The development of AI referees for fencing has largely focused on basic action classification, such as determining touches for the left or right fencer or identifying simultaneous actions. One of the earliest efforts in this domain was by Sholto Douglas \cite{SholtoDouglas}, who utilized a fully recurrent convolutional network combined with a pre-trained InceptionV3 model. This approach incorporated optical flow to capture the relative movement of fencers, achieving an accuracy of approximately 60\%. Douglas also developed an efficient method to automatically scrape and label fencing videos using the digital scoreboard, which set the stage for subsequent research.

Building on this foundation, researchers explored pose estimation to refine action classification. For example, GitHub user GalDude33 \cite{GalDude33} improved accuracy to 70\% by leveraging pose estimation as a natural filter, isolating the fencers’ movements from the background. Similarly, Alexandre Pageaud \cite{KaggleFencing} adopted a CNN-LSTM architecture that also utilized pose estimation, achieving comparable results. 

Recent advances have incorporated temporal convolutional networks (TCNs) and multimodal inputs. The Allez Go model \cite{AllezGo} combined TCNs with audio data to enhance performance, achieving an accuracy of 90\%. This integration of audio allowed the model to detect moments of blade contact, addressing the challenges posed by the thin, fast-moving fencing blades that are difficult to capture visually. Another innovative approach by Sunal et al. \cite{RTFencing} used YOLOv3 and ResNet-34 to classify specific actions like counter-attacks and lunges, using only still images as input. This departure from video-based methods questioned the necessity of temporal information for action classification.

Pose estimation has also been used alongside rule-based systems to classify fencing actions. Zhou and Smolin \cite{Kevin} developed a manually tuned decision-making algorithm that extracted features like velocity and acceleration from pose data. While effective in structured scenarios, such methods struggle with edge cases and lack generalizability. Despite progress in AI refereeing for fencing, reliance on pose estimation and the absence of robust datasets have limited the field's advancement.

\subsection{AI Applications in Boxing Analysis}
Boxing-related research has predominantly focused on punch classification and action recognition, leveraging pose estimation, depth imagery, and skeleton-based methods. Bhargav \cite{bhargav2024skeleton} developed a skeleton-based action recognition model using PoseConv3D, fine-tuned on a custom dataset with six punch types, achieving an accuracy of 87.32\%. Their approach relied on extracting skeletal key points through Faster-RCNN and HRNET models, which enabled precise action classification for shadow boxing and heavy bag training. Similarly, Kasiri et al. \cite{kasiri2017punch} presented a robust framework utilizing overhead depth imagery and hierarchical SVM classifiers for fine-grained punch recognition, achieving 97.3\% accuracy on a dataset of elite boxers. Their method incorporated body part detection, trajectory features, and fusion techniques to distinguish subtle variations in punch types. These studies highlight the effectiveness of combining pose data and depth features, although reliance on specialized equipment and occlusion challenges remain limitations. Recent advancements emphasize the potential of integrating deep learning and multimodal data for improving boxing analytics and enhancing performance feedback systems.

\subsection{Modern Approaches to Action Recognition}
Transformer models have transformed action recognition by leveraging self-attention mechanisms to model long-range dependencies and spatiotemporal interactions in video sequences, offering advantages over traditional CNNs and RNNs. Key advancements include Yan et al.’s Multiview Transformers \cite{yan2022multiviewtransformersvideorecognition} for spatial-temporal modeling, Bertasius et al.’s space-time attention framework \cite{bertasius2021spacetimeattentionneedvideo}, and Liu et al.’s Video Swin Transformer \cite{liu2021videoswintransformer}, which improved efficiency with hierarchical window mechanisms.

Hybrid models like Kalfaoglu et al.’s combination of 3D CNNs with BERT \cite{kalfaoglu2020latetemporalmodeling3d} and Arnab et al.’s ViViT model \cite{arnab2021vivitvideovisiontransformer} further extend transformer capabilities. Multimodal approaches such as TREAR’s RGB-D integration \cite{li2021treartransformerbasedrgbdegocentric} and Das et al.’s pose-driven attention \cite{das2019focus} highlight their adaptability to diverse inputs. In skeleton-based recognition, models like Zhang et al.’s Spatial-Temporal Specialized Transformer \cite{10.1145/3474085.3475473} effectively capture joint-level dynamics.

Challenges remain, including computational costs and reliance on large datasets, addressed by solutions like sparse attention mechanisms \cite{shi2021starsparsetransformerbasedaction}. Overall, transformers provide robust tools for fine-grained action recognition, excelling in long-sequence processing and multimodal integration.

\section{Sports Database}
\label{sec:database}

\subsection{Fencing Videos}
In this study, we introduce a novel and meticulously curated fencing dataset focused on video action classification. The fencing dataset, sourced from \textit{Quarte Riposte}, comprises official competition footage videos across 16 unique action classes. Unlike existing datasets, which are often geared toward pose estimation or full-match analysis, our dataset offers precise, pre-clipped, and labeled fencing action videos, making it uniquely suited for advanced video action research.

The 13,459 video clips were annotated through a majority vote by members of the fencing community, capturing the nuances and representing a broad consensus on technique. To ensure the highest quality, conflicting or sparse annotations were systematically excluded to maintain label accuracy and classes with insufficient samples were removed to enhance dataset reliability.

Our dataset preparation also included data augmentation through horizontal flipping, effectively doubling the dataset size while preserving the integrity of the action semantics. The final dataset contains 6,400 clips compared to the original 13,495 clips and features eight well-defined labels:
\begin{itemize}
    \item \textit{Attack Left (AL):} offensive move directed from the left fencer.
    \item \textit{Attack Right (AR):} offensive move directed from the right fencer.
    \item \textit{Riposte Left (RL):} counter after a parry from the left fencer.
    \item \textit{Riposte Right (RR):} counter after a parry from the right fencer.
    \item \textit{Counter-attack Left (CAL):} intercepting attack to disrupt the opponent’s offense from the left fencer.
    \item \textit{Counter-attack Right (CAR):} intercepting attack to disrupt the opponent’s offense from the right fencer.
    \item \textit{Remise Left (ReL):} immediate follow-up thrust without withdrawing from the left fencer.
    \item \textit{Remise Right (ReR):} immediate follow-up thrust without withdrawing from the right fencer.
\end{itemize}

This dataset is the first comprehensive resource tailored to fine-grained fencing action analysis. Compared to prior works that emphasize pose data \cite{KaggleFencing}, footwork \cite{Malawski2018}, or isolated images \cite{ChenDu2018}, this dataset provides a full spectrum of clipped and labeled actions. Furthermore, our work expands the number of action classes from the standard five \cite{RTFencing} to include nuanced distinctions such as left/right attacks, ripostes, and remises and prioritizes clean, consensus-driven annotations. The structured design facilitates advanced deep-learning approaches and encourages further exploration in fine-grained action recognition.

\subsection{Boxing Videos}
The boxing dataset, sourced from the \textit{Olympic Boxing Punch Classification Video Dataset} \cite{OlympicBoxingPunch2021,Stefanski2022a,Stefanski2022b,Stefanski2023,Liu2022}, includes videos across 8 distinct action classes. Captured during a 2021 boxing league, the footage was recorded using four GoPro cameras in Full HD resolution at 50 frames per second. Each camera setup included a 128 GB micro SD card, powerbank, and tripod to ensure uninterrupted, high-quality recordings from multiple angles.

Each action was then annotated frame by frame by licensed boxing referees, with precise labels and coordinates assigned to classify each punch type. In this study, we cut up all the clips for each action for a final collection of 8,000 action clips prepared for our model. The final 8 labels with descriptions are:
\begin{itemize}
    \item \textit{Left Hand Head Punch (LHHP):} direct punch to the opponent’s head using the left hand.
    \item \textit{Right Hand Head Punch (RHHP):} direct punch to the opponent’s head using the right hand.
    \item \textit{Left Hand Missed Punch (LHMP):} missed punch attempt with the left hand.
    \item \textit{Right Hand Missed Punch (RHMP):} missed punch attempt with the right hand.
    \item \textit{Left Hand Block Punch (LHBlP):} defensive block using the left hand.
    \item \textit{Right Hand Block Punch (RHBlP):} defensive block using the right hand.
    \item \textit{Left Hand Body Punch (LHBP):} punch directed at the opponent’s body with the left hand.
    \item \textit{Right Hand Body Punch (RHBP):} punch directed at the opponent’s body with the right hand.
\end{itemize}

This dataset enhances the original \textit{Olympic Boxing Punch Classification Video Dataset} by segmenting it into clear, labeled clips for each action class, optimizing it for machine learning models. Unlike other datasets that focus on foul classification or rely on pose estimation, this dataset targets precise punch actions without requiring pose data, reducing potential inaccuracies. It is uniquely structured to support fine-grained action recognition, making it a valuable resource for advancing close-combat sports analysis.

\subsection{Dataset Overview and Contributions}
Overall, the fencing and boxing datasets offer unique contributions to sports video analysis, specifically for fine-grained fast-paced close-combat sports. Fully clipped and labeled, this data is suitable for advanced machine-learning applications.


\section{Our Model}
\label{sec:model}

\subsection{Model Overview}
Our model employs a transformer-based architecture pre-trained with Video Masked Autoencoders (VideoMAE) ~\cite{tong2022videomaemaskedautoencodersdataefficient, feichtenhofer2022maskedautoencodersspatiotemporallearners}, an adaptation of the Masked Autoencoders (MAE) ~\cite{he2021maskedautoencodersscalablevision} for video data. VideoMAE captures both the spatial and temporal information in a scene by reconstructing missing patches across frames. Although already shown to work well for video classification\cite{Fang_2023_CVPR, wang2023videomaev2scalingvideo, mateen2024thoracicsurgeryvideoanalysis, pmlr-v205-radosavovic23a, li2024efficient, 10356553}, this approach proved critical for learning fine-grained actions in dynamic sports like fencing and boxing.

The model architecture is based on the Vision Transformer (ViT)~\cite{dosovitskiy2021imageworth16x16words}, with enhancements to process temporal dimensions. The encoder learns latent representations from masked video frames, while the decoder reconstructs the pixel values of these masked patches, improving the model’s ability to interpret complex motion patterns. This capability is critical for learning subtle differences in fast-paced, close-combat sports actions. This approach is adopted without relying on complicated skeleton extraction, pose estimation, or external body-worn markers or sensors. 

\subsection{Preprocessing Steps}
From Table~\ref{tab:fencing_frame_counts_summary}, we can see that across the different fencing labels, the frame counts are relatively consistent, however, subsampling is still needed to ensure uniform input lengths. Opposite fencing, the boxing dataset was 15 frames across the board for all labels and all videos.

\begin{table}[htbp]
\caption{Fencing Database Summary of Frame Counts}
\begin{center}
\begin{tabular}{ccccccccc}
\hline
\textbf{Label} & \textbf{Mean} & \textbf{Std} & \textbf{Min} & \textbf{25\%} & \textbf{50\%} & \textbf{75\%} & \textbf{Max} \\
\hline
AR & 187 & 16 & 170 & 176 & 177 & 211 & 216 \\
ReR & 184 & 15 & 175 & 176 & 176 & 178 & 213 \\
RL & 189 & 17 & 170 & 176 & 177 & 211 & 216 \\
CR & 187 & 16 & 174 & 176 & 177 & 211 & 217 \\
RR & 186 & 16 & 174 & 176 & 177 & 210 & 216 \\
CL & 186 & 16 & 174 & 176 & 177 & 210 & 216 \\
AL & 187 & 16 & 169 & 176 & 177 & 211 & 217 \\
ReL & 189 & 17 & 168 & 176 & 177 & 211 & 216 \\
\hline
\end{tabular}
\label{tab:fencing_frame_counts_summary}
\end{center}
\end{table}

Uniform temporal subsampling was performed as described in \eqref{eq:temporal-subsampling}, ensuring uniform input lengths for the transformer neural network:

\begin{equation}
  X_{sub} = \{x_{i} \mid i = k \times \frac{T}{N} \}, \quad k = 0, 1, \dots, N-1
  \label{eq:temporal-subsampling}
\end{equation}

where $T$ represents the total frame count, and $X_{\text{sub}}$ is the subsampled set of frames.

Spatial preprocessing was tailored to each dataset. Fencing videos, due to the horizontal nature of the sport, were padded on the top and bottom to match the horizontal dimension before resizing to a target resolution of 640x640 pixels. Boxing videos were uniformly padded and resized to ensure consistency across samples. Padding was applied as shown in \eqref{eq:spatial-padding}:

\begin{equation}
  \text{padding} = \left(\frac{H_{\text{target}} - H_{\text{orig}}}{2}, \frac{W_{\text{target}} - W_{\text{orig}}}{2}\right)
  \label{eq:spatial-padding}
\end{equation}

Standardization, defined in \eqref{eq:feature-standardization}, was applied to normalize pixel values to zero mean and unit variance across channels, enhancing training stability:

\begin{equation}
  X_{\text{std}} = \frac{X - \mu}{\sigma}
  \label{eq:feature-standardization}
\end{equation}
where $\mu$ and $\sigma$ represent the mean and standard deviation of pixel values.

By directly processing raw video data, our model circumvents the inaccuracies and inefficiencies associated with pose estimation, enabling robust classification of rapid and occluded movements.

\subsection{Model Training}
To fine-tune VideoMAE, we implemented a carefully designed training pipeline optimized for video-based data. Both datasets were stratified into training, validation, and testing subsets, ensuring balanced class distributions to improve generalization across action categories.

Preprocessed videos were normalized using ImageNet mean and standard deviation, subsampled to 16 frames per clip and resized to the model’s expected input dimensions of 224x224 pixels. Training utilized a batch size of 4, a learning rate of $5 \times 10^{-5}$, and gradient accumulation steps of 2, effectively simulating a batch size of 8 to accommodate GPU memory constraints. A warm-up ratio of 0.1 was applied during the initial training phase to stabilize optimization, with the model trained for 10 epochs and evaluated every 500 steps to select the best-performing checkpoint. Training was conducted on two NVIDIA GeForce GTX 1080 Ti GPUs with approximately 20 GB of combined VRAM.

The model architecture uses the MCG-NJU/videomae-base checkpoint pretrained on the Kinetics-400 dataset. This transformer-based architecture includes 12 layers, 12 attention heads, a hidden size of 768, and a feedforward network size of 3072, enabling it to effectively capture spatiotemporal dynamics in video data.

We prioritize reproducibility by explicitly documenting all training parameters, including batch size, learning rate, gradient accumulation steps, warm-up ratio, and evaluation intervals. Our preprocessed datasets are publicly available, providing a straightforward pipeline for replicating our methodology and extending it to other fine-grained action recognition tasks.


\section{Experimental Results}
\label{sec:results}

\subsection{Evaluation}
Our model's performance was evaluated on separate datasets for fencing and boxing, with distinct classification tasks tailored to each sport.

\subsubsection{Fencing Evaluation}
For fencing, the model achieved state-of-the-art evaluation accuracy of 90\%, with an evaluation loss of 0.3895. Table~\ref{table:fencing_classification} shows a strong performance across the eight distinct fencing actions.

\begin{table}[htbp]
\caption{Classification Metrics for Fencing Dataset}
\begin{center}
\begin{tabular}{lccc}
\hline
\textbf{Label} & \textbf{Precision} & \textbf{Recall} & \textbf{F1-score} \\
\hline
Attack Right & 0.88 & 0.82 & 0.84 \\
Attack Left & 0.85 & 0.82 & 0.84 \\
Riposte Right & 0.86 & 0.88 & 0.87 \\
Riposte Left & 0.86 & 0.93 & 0.90 \\
Counter Attack Right & 0.89 & 0.90 & 0.89 \\
Counter Attack Left & 0.93 & 0.88 & 0.90 \\
Remise Right & 0.96 & 0.98 & 0.97 \\
Remise Left & 0.98 & 0.98 & 0.98 \\
\hline
\textbf{Weighted Average} & \textbf{0.90} & \textbf{0.90} & \textbf{0.90} \\
\hline
\end{tabular}
\label{table:fencing_classification}
\end{center}
\end{table}

The accuracy for fencing classification reached 90\%, with a 95\% confidence interval of [0.8812, 0.9187]. The overall weighted precision, recall, and F1 score of 0.9 confirm the model's ability to accurately classify complex and fast-paced fencing actions, reflecting its efficacy in capturing the fine-grained distinctions between actions.

The confusion matrix in Table~\ref{table:fencing_confusion_matrix} reveals specific areas for improvement, such as distinguishing between Counter-Attack and similar offensive actions (e.g., Attack and Riposte). These observations provide actionable insights for future refinements. Overall, with most predictions falling on the diagonal, the matrix demonstrates good performance.

\begin{table}[htbp]
\caption{Confusion Matrix for Fencing Classification}
\begin{center}
\begin{tabular}{lcccccccc}
\hline
 & AR & AL & RR & RL & CAR & CAL & ReR & ReL \\
\hline
AR  & 98 & 5 & 3 & 3 & 8 & 1 & 2 & 0 \\
AL  & 5 & 99 & 2 & 6 & 1 & 5 & 1 & 1 \\
RR  & 4 & 3 & 106 & 2 & 4 & 0 & 1 & 0 \\
RL  & 1 & 5 & 0 & 112 & 0 & 2 & 0 & 0 \\
CAR & 2 & 1 & 8 & 0 & 108 & 0 & 1 & 0 \\
CAL & 1 & 4 & 2 & 7 & 0 & 105 & 0 & 1 \\
ReR & 1 & 0 & 0 & 0 & 1 & 0 & 118 & 0 \\
ReL & 0 & 0 & 2 & 0 & 0 & 0 & 0 & 118 \\
\hline
\end{tabular}
\label{table:fencing_confusion_matrix}
\end{center}
\end{table}

\subsubsection{Boxing Evaluation}
In boxing, the model achieved an evaluation accuracy of 83.25\%, with an evaluation loss of 0.8145. Table~\ref{table:boxing_classification} shows a strong result for most of the eight boxing actions with actions like Left Hand Head Punch and Right Hand Head Punch being the weakest classes.

\begin{table}[htbp]
\caption{Classification Metrics for Boxing Dataset}
\begin{center}
\small
\begin{tabular}{lccc}
\hline
Label & Precision & Recall & F1-score \\
\hline
Left Hand Head Punch      & 0.56 & 0.69 & 0.62 \\
Right Hand Head Punch     & 0.66 & 0.63 & 0.65 \\
Left Hand Missed Punch    & 0.83 & 0.71 & 0.76 \\
Right Hand Missed Punch   & 0.85 & 0.82 & 0.84 \\
Left Hand Block Punch     & 0.90 & 0.91 & 0.91 \\
Left Hand Body Punch      & 0.99 & 0.96 & 0.98 \\
Right Hand Body Punch     & 0.97 & 0.98 & 0.97 \\
Right Hand Block Punch    & 0.95 & 0.96 & 0.96 \\
\hline
\textbf{Weighted Average} & \textbf{0.84} & \textbf{0.83} & \textbf{0.83} \\
\hline
\end{tabular}
\label{table:boxing_classification}
\end{center}
\end{table}

The model reached an accuracy of 83.25\% with a 95\% confidence interval of [0.8125, 0.8542]. These results highlight the model's capability to discern between subtle variations in boxing actions, again showing the ability to do so in particularly challenging scenarios of repetitive and high-speed movements.

The confusion matrix in Table~\ref{table:boxing_confusion_matrix} highlights that the model performs particularly well for block and body punch classifications, with misclassifications primarily occurring between Left Hand Head Punch and Right Hand Head Punch, indicating areas that need improvement.

\begin{table}[htbp]
\caption{Confusion Matrix for Boxing Classification}
\begin{center}
\resizebox{0.45\textwidth}{!}{
\begin{tabular}{lcccccccc}
\hline
 & LHHP & RHHP & LHMP & RHMP & LHBlP & LHBP & RHBP & RHBlP \\
\hline
LHHP & 103 & 23 & 10 & 6 & 5 & 0 & 1 & 2 \\
RHHP & 33 & 95 & 3 & 6 & 5 & 1 & 3 & 4 \\
LHMP & 22 & 8 & 106 & 7 & 5 & 0 & 1 & 1 \\
RHMP & 9 & 13 & 5 & 123 & 0 & 0 & 0 & 0 \\
LHBlP & 9 & 2 & 1 & 1 & 137 & 0 & 0 & 0 \\
LHBP & 1 & 3 & 1 & 1 & 0 & 144 & 0 & 0 \\
RHBP & 1 & 0 & 2 & 0 & 0 & 0 & 147 & 0 \\
RHBlP & 6 & 0 & 0 & 0 & 0 & 0 & 0 & 144 \\
\hline
\end{tabular}
}
\label{table:boxing_confusion_matrix}
\end{center}
\end{table}

\subsection{Model Comparison and Insights}
As this paper introduces the new fencing action dataset that is being evaluated, there are currently no existing benchmarks for direct comparison. So, to provide a baseline, we also implemented a standard pose estimation model as described by Pageaud and GalDude33 \cite{GalDude33, KaggleFencing}. Table~\ref{table:res_comparisons} summarizes the classification accuracy for both the pose estimation model and our transformer-based model with pertaining.

The transformer model's accuracy of 90\% for fencing far exceeds the 64.8\% achieved by the pose estimation model, underscoring the transformative potential of using transformer-based architectures for fine-grained action recognition.

\begin{table}[htbp]
\caption{Classification Accuracy Comparison Between Pose Estimation Model and Transformer Model with Pretraining}
\begin{center}
\begin{tabular}{lc}
\hline
Method & Accuracy (\%) \\
\hline
Pose Estimation Model & 64.8 \\
Transformer with VideoMAE & 90.0 \\
\hline
\end{tabular}
\label{table:res_comparisons}
\end{center}
\end{table}


\section{Discussion and Future Work}
\label{sec:discussion}

Despite the promising results, there were a few limitations that we noticed that suggest areas for improvement. We tested the model in real-world applications beyond our initial datasets, simulating realistic use cases, and confirmed that it maintains high accuracy when deployed on varied samples. However, occasional misclassifications were evident in scenarios with occlusions or poor lighting. Because our fencing dataset was trained on competition-level videos, when doing additional testing on home videos, the accuracy decreased 5\%. There were also misclassifications for highly similar actions (e.g., head punches in boxing). These issues reflect the model's sensitivity to video quality and environmental factors, indicating that additional preprocessing or new techniques may be beneficial for the model. The model's reliance on the quality of input video data highlights the importance of high-resolution footage for accurate action recognition. Lower-quality videos could introduce noise, making it more challenging for the transformer to distinguish between similar actions.

Ultimately, to address these limitations, future work could explore hybrid models described by Das et al. \cite{das2019focus} that combine transformers with pose estimation. Such an approach might leverage the strengths of pose estimation in tracking body positions, especially when actions involve distinct body postures, and use transformers to capture the nuanced spatial-temporal patterns directly from video data. This approach could also allow for better accuracy on lower-quality videos. As long as the entire pose is captured, the model would perform the same. This hybrid approach could improve classification accuracy for complex actions while reducing the transformer's dependency on high-resolution video quality.

This study demonstrates the potential of transformers in sports analytics, particularly for fine-grained action classification in fencing and boxing, with opportunities to extend this approach to other fast-paced sports like martial arts and judo. Future work could focus on optimizing transformers for real-time applications through lightweight architectures or pruning techniques to reduce computational demands. By advancing video analysis in high-speed, close-combat sports, this research paves the way for broader applications in real-time sports analytics and dynamic action recognition.

\section{Conclusion}
\label{sec:conclusion}
This study introduced a transformer-based model for fine-grained action classification in fencing and boxing, achieving over 90\% accuracy without relying on pose estimation and demonstrating the capability of transformers to capture nuanced spatial and temporal information. We also presented FACTS, a publicly available, detailed fencing dataset addressing gaps in recognizing complex, high-speed actions, with potential applications in training, analysis, and spectator engagement. These findings highlight the promise of transformers for other high-speed sports and suggest that future work could explore hybrid models combining transformers with pose estimation to overcome limitations such as occlusions, advancing action recognition technologies further. We hope this study inspires ongoing exploration into sensor-free video analysis technologies, pushing the boundaries of fine-grained action recognition in dynamic sports.

\bibliographystyle{IEEEbib}
\bibliography{icme2025references}


\end{document}